\documentclass[preprint,12pt]{elsarticle}

\usepackage{amsmath,amssymb,amsthm}
\usepackage{booktabs}
\usepackage{multirow}
\usepackage{rotating}
\usepackage{graphicx}
\usepackage{xcolor}
\usepackage{algorithm}
\usepackage{algpseudocode}
\usepackage{url}
\usepackage{hyperref}
\usepackage{array}

\biboptions{numbers,sort&compress}

\hypersetup{colorlinks=true, linkcolor=blue, citecolor=blue, urlcolor=blue}

\newcommand{\proposed}{\textbf{FedTGNN-SS}}

\begin{document}

\begin{frontmatter}

\title{Federated Semi-Supervised Graph Neural Networks with Prototype-Guided Pseudo-Labeling for Privacy-Preserving Gestational Diabetes Mellitus Prediction}

\author[aff1]{G.~Victor~Daniel\corref{cor1}}
\ead{victordanielai@anurag.edu.in}

\cortext[cor1]{Corresponding author.}

\affiliation[aff1]{
  organization={School of Engineering, Anurag University},
  addressline={Hyderabad, Telangana},
  country={India}
}


\begin{abstract}
Gestational Diabetes Mellitus (GDM) is a common pregnancy complication that requires early risk prediction to reduce maternal and fetal complications. However, clinical machine learning is often limited by scarce labeled electronic health records (EHRs) and strict privacy constraints across hospitals. To address these challenges, this paper proposes FedTGNN-SS, a privacy-preserving federated semi-supervised framework for clinical tabular EHR data. Each hospital constructs a local patient similarity graph and trains a topology-adaptive graph neural network (GNN). The framework leverages unlabeled data using prototype-guided pseudo-labeling, adaptive graph refinement, and clinical-aware consistency augmentation. To preserve privacy, only class-level prototype embeddings are exchanged between institutions instead of patient-level records. Experiments were conducted on three diabetes-related datasets under 10\%--80\% missing labels per silo. FedTGNN-SS achieved 56 statistically significant improvements ($p < 0.05$) over 11 federated baselines and maintained strong performance under severe label scarcity, achieving AUROC scores of 0.8037 on the Pima dataset and 0.9634 on the Early Stage dataset with 80\% missing labels. The results demonstrate that FedTGNN-SS is an effective and privacy-aware framework for semi-supervised clinical risk prediction in distributed healthcare environments.
\end{abstract}

\begin{keyword}
Gestational Diabetes Mellitus \sep Federated Learning \sep Graph Neural Networks \\ 
\sep Semi-Supervised Learning \sep Pseudo-Labeling \sep Privacy-Preserving \\
\sep Patient Similarity Graph \sep Label Scarcity \sep Clinical Decision Support \sep EHR
\end{keyword}

\end{frontmatter}


\section{Introduction}
\label{sec:intro}

Gestational diabetes mellitus (GDM) is reported to affect approximately 14–17\% of pregnancies worldwide ~\cite{IDF2021}. It is associated with increased risks of macrosomia, preeclampsia, and cesarean delivery; at the same time, it is linked to a higher likelihood of long-term progression to type 2 diabetes in both the mother and the child ~\cite{Vounzoulaki2020}. Early and accurate risk stratification enables timely clinical intervention and motivates an increase in work on machine learning (ML) prediction of GDM onset ~\cite{Kavakiotis2017,Naz2020,Zou2018}.
Despite promising results in controlled research settings,
two fundamental barriers prevent real-world deployment of
ML systems for GDM prediction.

The first barrier is  Label Scarcity. A confirmed diagnosis of gestational diabetes mellitus (GDM) requires the oral glucose tolerance test (OGTT) at 24–28 weeks of gestation.  This test is resource-intensive; as a result, it is not universally administered in routine antenatal care. Becasue of this, a large proportion of patient records – often 70–80\% – lack
confirmed GDM labels ~\cite{WHO_GDM_2013}. Supervised ML methods that
require fully labeled datasets fail catastrophically under such
annotation scarcity. Semi-supervised learning (SSL) provides a principled approach by exploiting unlabeled records as an additional training signal. At the same time, existing SSL methods for clinical data are affected by unreliable pseudo-label assignment under conditions of high data scarcity and class imbalance~\cite{Lee2013,Sohn2020}.

The second barrier is data privacy. Patient EHR data are legally
protected under HIPAA (USA), GDPR (EU), and equivalent national legislation. Centralizing patient records from multiple hospitals for model training is therefore not feasible without formal data-sharing agreements; in many cases, such agreements are logistically prohibitive at scale. Federated learning (FL) offers an
 alternative that preserves privacy by training models locally in each institution and sharing only model weights ~\cite{McMahan2017}.  However, existing federated learning (FL) methods for clinical tabular data are predominantly supervised; as a result, they do not take advantage of the large pool of unlabeled patient records available at each institution.

\textbf{The Unaddressed Combination.}
No existing method simultaneously addresses both barriers for clinical tabular
EHR data. Federated GNN methods (FedSage+~\cite{FedSage2021}, FedGL~\cite{FedGL2023})
require cross-silo graph edges -- a direct privacy violation in healthcare.
Federated SSL methods (FedMatch~\cite{Jeong2021}, RSCFed~\cite{Liang2022}) have
no graph structure and are evaluated exclusively on image benchmarks. A comprehensive
literature search across 2017--2026 confirms that the combination of
\emph{federated learning + semi-supervised GNN + clinical tabular data} has not
been addressed.

\textbf{Proposed Framework: FedTGNN-SS.}
We propose \proposed{}, a framework that addresses both barriers simultaneously.
Each hospital constructs a local patient similarity graph and trains a hybrid
GCN--GraphSAGE encoder locally. Four novel components distinguish \proposed{}
from all prior work:

\begin{enumerate}
  \item \textbf{Prototype-Guided Pseudo-Labeling (PG-PL)} assigns pseudo-labels to unlabeled patient nodes only when (a) softmax confidence exceeds a decaying threshold $\tau_t$, (b) the predicted class matches the
        nearest class centroid (prototype) in the embedding space, and (c) a
        majority of graph neighbors agree. This triple gate dramatically reduces
        error accumulation compared to confidence-only thresholding used in
        GRAND~\cite{GRAND2020} and FixMatch~\cite{Sohn2020}.

        \item \textbf{Adaptive Graph Refinement (AGR)} periodically reconstructs the
local k-NN graph every K training epochs using the evolving GNN
embeddings rather than the original clinical features. As a result of this, the
graph progressively reflects more clinically meaningful patient relation-
ships. This mitigates the limitation of static graph structures in exist-
ing federated GNN baselines.

         \item \textbf{Clinical-Aware Augmentation (CAA)} applies consistency regu-
larization by introducing Gaussian noise only to continuous clinical
features (e.g., BMI, OGTT, systolic blood pressure). At the same time it preserves binary and categorical features (e.g., PCOS, family history, parity).
This results in physiologically plausible augmentations and mitigates
the semantic distortion caused by random feature dropout methods
such as GRAND on tabular data.

\item \textbf{Privacy-Safe Prototype Sharing (PS)} enables each silo to share class-level prototype vectors, defined as embedding centroids of labeled nodes, after every federation round instead of patient features or graph edges. The server aggregates these prototypes globally, and each silo uses the aggregated representation to improve local pseudo-labeling. This mechanism facilitates cross-silo pseudo-label refinement without requiring patient data transfer.

\end{enumerate}

\begin{figure}[!t]
  \centering
  \includegraphics[width=\linewidth,height=0.75\textheight,keepaspectratio]{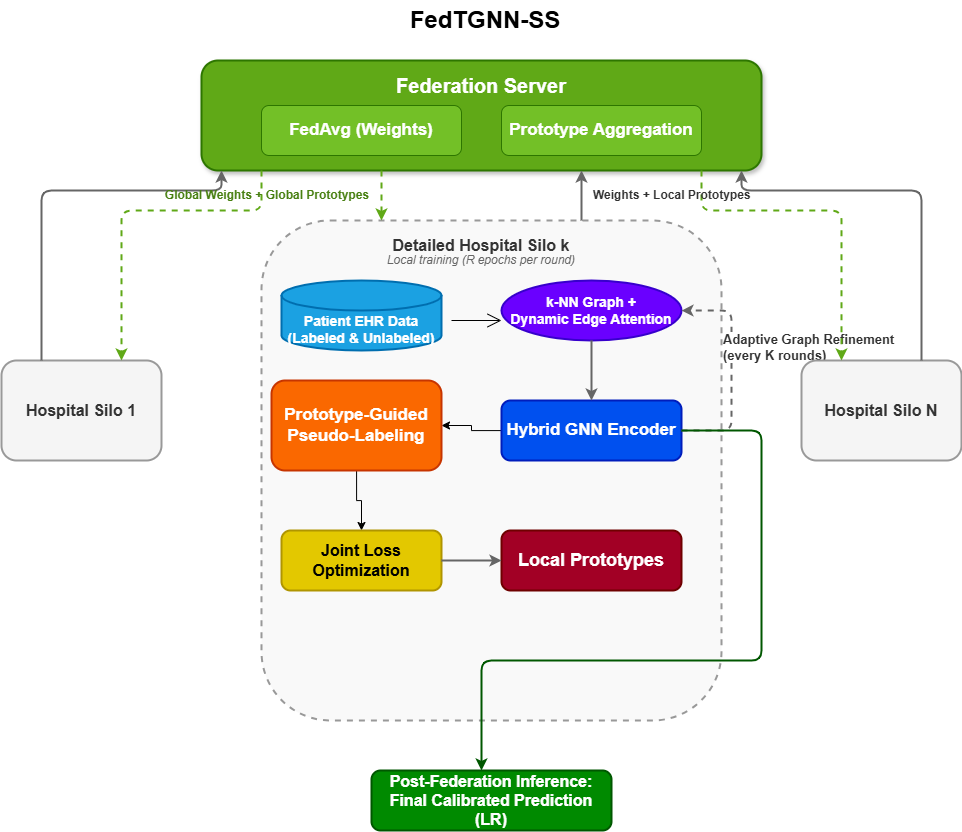}
  \caption{Overview of the proposed \proposed{} framework. Each hospital trains a local GNN on its patient similarity graph with (i) prototype-guided pseudo-labeling, (ii) adaptive graph refinement, and (iii) clinical-aware augmentation; only class-level prototypes are shared with the server for privacy-safe aggregation across rounds.}
  \label{fig:framework}
\end{figure}

\textbf{Contributions.}
The specific contributions of this paper are:

\begin{enumerate}
  \item \proposed{}: the first federated semi-supervised GNN framework for
        clinical tabular EHR data, confirmed to fill a gap in the 2017--2026
        literature (Section~\ref{sec:related}).

  \item Prototype-guided pseudo-labeling combined with adaptive graph
        refinement which  is a novel combination that reduces pseudo-label error
        accumulation and overcomes static graph limitations in federated settings
        (Section~\ref{sec:method}).

  \item Privacy-safe prototype sharing: It is  a mechanism enabling cross-silo
        pseudo-label improvement without transmitting patient features or
        graph connectivity (Section~\ref{sec:method}).

  \item Comprehensive empirical evaluation across three clinical datasets,
        five scarcity levels, 5-fold cross-validation, and 11 baselines with
        Wilcoxon signed-rank statistical testing (Sections~\ref{sec:exp}
        and~\ref{sec:results}).
\end{enumerate}

The remainder of this paper is organised as follows.
Section~\ref{sec:related} surveys related work and identifies specific
limitations. Section~\ref{sec:method} presents the \proposed{} framework.
Section~\ref{sec:exp} describes experimental setup. Section~\ref{sec:results}
presents results and statistical analysis. Section~\ref{sec:conclusion}
concludes with future directions.


\section{Related Work}
\label{sec:related}

\subsection{GDM and Diabetes Prediction with Machine Learning}

Classical supervised ML methods -- Logistic Regression, Random Forest, SVM,
and XGBoost -- have been extensively applied to GDM and diabetes prediction
on single-site tabular datasets~\cite{Kavakiotis2017,Zou2018,Naz2020,Darrar2017}.
A meta-analysis of GDM prediction models reports a pooled AUROC of 0.849 across
studies using these methods~\cite{GDM_BMC2024}. A limitation common to all these
works is the requirement for fully labeled datasets: performance degrades sharply
as the labeled fraction decreases. Furthermore, all evaluations are conducted
in centralised single-site settings and cannot be directly applied across hospital
boundaries due to data governance constraints. Privacy-preserving approaches using
federated learning for diabetes prediction have been explored (e.g.,
\cite{Tang2024Canada, FedEnTrust2026}), but these works use only supervised
FedAvg with tabular classifiers and do not address label scarcity or graph structure.

\subsection{Graph Neural Networks for Semi-Supervised Learning}

Graph Convolutional Networks (GCN)~\cite{Kipf2017} and GraphSAGE~\cite{Hamilton2017}
introduced transductive and inductive GNN architectures for semi-supervised node
classification. Both rely on static graphs with fixed edge weights and have no
pseudo-labeling mechanism. GRAND~\cite{GRAND2020} addresses over-smoothing via
random feature dropout for data augmentation and consistency regularization.
However, (i) feature dropout is not semantically meaningful for clinical tabular
data where each feature carries specific physiological meaning (e.g., dropping OGTT
or BMI destroys clinical signal rather than creating a valid augmentation), and
(ii) GRAND treats all unlabeled nodes equally in the consistency loss regardless
of prediction confidence, leading to noise accumulation under high label scarcity.
FixMatch~\cite{Sohn2020} (adapted to graphs) uses a fixed confidence threshold
$\tau = 0.95$ with no prototype validation or decay schedule. GraphMix~\cite{Verma2021}
applies manifold mixup regularization but produces clinically implausible
interpolated patient representations. None of these methods operate in a federated
setting.

\subsection{Federated Learning}

McMahan et al.~\cite{McMahan2017} proposed FedAvg, the foundational federated
averaging algorithm. FedProx~\cite{FedProx2020} added a proximal regularization
term to limit client drift under non-IID data. SCAFFOLD~\cite{SCAFFOLD2020}
uses control variates to correct client drift but at twice the communication
cost. FedBN~\cite{FedBN2021} maintains local batch normalization statistics to
address feature shift, though it is primarily designed for image models with
BatchNorm layers and does not handle label scarcity. All of these methods are
exclusively supervised -- they cannot exploit unlabeled patient records.

\subsection{Federated Graph Neural Networks}

FedGraphNN~\cite{FedGraphNN2021} provides a benchmark for federated GNN training
and reports a key finding: GNN models that outperform others in centralized
settings do not maintain this advantage in federated non-IID settings. FedSage+
\cite{FedSage2021} generates missing cross-silo neighbors to recover graph
connectivity, but this requires sharing node embeddings across silos -- a
potential privacy violation in healthcare, where embedding inversion attacks
can recover patient features~\cite{Fredrikson2015}. SpreadGNN~\cite{SpreadGNN2022}
operates in a serverless federated setting for molecular graphs, which does not
transfer to patient EHR tabular data. FedGL~\cite{FedGL2023} discovers global
pseudo-graphs by exchanging prediction logits, introducing noise into graph
structure. A 2026 federated GNN work for clinical event prediction~\cite{MCNGNN2026}
requires blockchain coordination (adding $>$200ms latency) and has no SSL
component. No existing federated GNN method operates on clinical tabular data with
semi-supervised learning.

\subsection{Federated Semi-Supervised Learning}

FedMatch~\cite{Jeong2021} introduces inter-client consistency regularization for
federated SSL, assuming that silo models should produce similar predictions on
unlabeled data -- an assumption that fails when hospitals serve legitimately
different patient populations. RSCFed~\cite{Liang2022} uses random sub-consensus
aggregation with distance-reweighted model averaging, but is evaluated only on
image benchmarks (CIFAR-10, SVHN). FedSemi~\cite{FedSemi2020} employs a
teacher-student framework with adaptive layer selection but doubles per-client
computation. MedFedProto~\cite{MedFedProto2025} applies federated prototypical
learning to medical images, not clinical tabular data. CCWFed~\cite{CCWFed2025}
addresses class imbalance in federated SSL via calibration weighting but has no
graph structure and is evaluated on imaging tasks. None of these methods
incorporate patient similarity graphs.

\subsection{Positioning of \proposed{}}

Table~\ref{tab:related} summarises the positioning of \proposed{} against key
prior works along five axes: federated training, semi-supervised SSL, graph
structure, clinical tabular data, and label imbalance handling. \proposed{} is
the only method satisfying all five criteria.

\begin{table}[!t]
\caption{Comparison of \proposed{} with representative prior works.
  \checkmark = supported; \texttimes = not supported; $\sim$ = partial.}
\label{tab:related}
\centering
\small
\setlength{\tabcolsep}{3pt}
\begin{tabular}{lccccc}
\toprule
\textbf{Method} & \textbf{Fed.} & \textbf{SSL} & \textbf{Graph}
  & \textbf{Tabular} & \textbf{Imbalance} \\
\midrule
FedAvg~\cite{McMahan2017}         & \checkmark & \texttimes & \texttimes & \checkmark & \texttimes \\
FedProx~\cite{FedProx2020}        & \checkmark & \texttimes & \texttimes & \checkmark & \texttimes \\
FedMatch~\cite{Jeong2021}         & \checkmark & \checkmark & \texttimes & \texttimes & \texttimes \\
RSCFed~\cite{Liang2022}           & \checkmark & \checkmark & \texttimes & \texttimes & \texttimes \\
FedSage+~\cite{FedSage2021}       & \checkmark & \texttimes & $\sim$     & \texttimes & \texttimes \\
GCN~\cite{Kipf2017}               & \texttimes & \checkmark & \checkmark & \checkmark & \texttimes \\
GRAND~\cite{GRAND2020}            & \texttimes & \checkmark & \checkmark & $\sim$     & \texttimes \\
FixMatch~\cite{Sohn2020}          & \texttimes & \checkmark & $\sim$     & \texttimes & \texttimes \\
MedFedProto~\cite{MedFedProto2025}& \checkmark & \checkmark & \texttimes & \texttimes & \texttimes \\
\midrule
\textbf{FedTGNN-SS (Ours)}        & \checkmark & \checkmark & \checkmark & \checkmark & \checkmark \\
\bottomrule
\end{tabular}
\end{table}


\section{Methodology}
\label{sec:method}

The \proposed{} framework consists of three stages executed at each federation
round: (1) local graph construction and GNN training within each hospital silo,
(2) federated model aggregation via FedAvg with a proximal constraint, and
(3) privacy-safe prototype exchange across silos. We describe each component in
detail below.

\subsection{Problem Formulation}

Let $\mathcal{H} = \{H_1, H_2, \ldots, H_S\}$ denote $S$ hospital silos.
Each silo $H_k$ holds a private dataset $\mathcal{D}_k = \{(\mathbf{x}_i, y_i)\}
\cup \{\mathbf{x}_j\}$, where $\mathbf{x}_i \in \mathbb{R}^d$ is the clinical
feature vector of patient $i$, $y_i \in \{0, 1\}$ is the GDM label (available for
a labeled subset $\mathcal{L}_k$), and $\{\mathbf{x}_j\}$ are unlabeled records
in $\mathcal{U}_k$. Let $\rho = |\mathcal{U}_k| / (|\mathcal{L}_k| + |\mathcal{U}_k|)$
denote the label scarcity ratio ($\rho \in \{0.1, 0.3, 0.5, 0.7, 0.8\}$ in our
experiments). Patient features are never shared across silos.

The objective is to learn a GDM prediction function $f: \mathbb{R}^d \to [0,1]$
that: (i) is trained without centralising patient data, (ii) leverages unlabeled
records in each silo, and (iii) produces well-calibrated GDM risk probability
estimates.

\subsection{Local Patient Similarity Graph}

Within each silo $H_k$, we construct a patient similarity graph
$\mathcal{G}_k = (\mathcal{V}_k, \mathcal{E}_k, \mathbf{W}_k)$ where nodes
$\mathcal{V}_k$ represent patients and edges $\mathcal{E}_k$ connect the $k$
most similar patients for each node. Edge weights are computed via a Gaussian
kernel on Euclidean feature distances:
\begin{equation}
  w_{ij} = \exp\!\left(-\frac{\|\mathbf{x}_i - \mathbf{x}_j\|^2}{2\sigma^2}\right),
  \quad \sigma = \text{median}_{(i,j)} \|\mathbf{x}_i - \mathbf{x}_j\|
\end{equation}
No cross-silo edges are created. Each silo's graph is fully local, ensuring
that patient connectivity information never leaves the institution.

\subsection{FedTGNN-SS Architecture}

\subsubsection{Dynamic Edge Attention}
Standard $k$-NN graphs assign edge weights based solely on raw feature distances
computed before training. As the GNN learns richer representations, these static
weights become suboptimal. We introduce a \emph{dynamic edge attention} module
that learns per-edge scalar weights from concatenated node feature pairs:
\begin{equation}
  a_{ij} = \sigma\!\left(\text{MLP}([\mathbf{x}_i \,\|\, \mathbf{x}_j])\right),
  \quad a_{ij} \in (0, 1)
\end{equation}
where $\sigma$ denotes the sigmoid function and $[\cdot\|\cdot]$ is concatenation.
The MLP consists of two linear layers (input: $2d$, hidden: 64, output: 1).
These attention weights gate the static Gaussian weights during message passing.

\subsubsection{Hybrid GCN--GraphSAGE Encoder}
We employ a two-layer hybrid encoder combining GCN~\cite{Kipf2017} and
GraphSAGE~\cite{Hamilton2017} with a learnable fusion parameter $\alpha$:
\begin{align}
  \mathbf{H}^{(1)} &= \alpha \cdot \text{GCN}(\mathbf{X}, \mathcal{E}, \mathbf{a})
    + (1-\alpha) \cdot \text{SAGE}(\mathbf{X}, \mathcal{E}) \\
  \mathbf{H}^{(2)} &= \alpha \cdot \text{GCN}(\mathbf{H}^{(1)}, \mathcal{E}, \mathbf{a})
    + (1-\alpha) \cdot \text{SAGE}(\mathbf{H}^{(1)}, \mathcal{E})
\end{align}
GCN captures spectral graph structure through normalized Laplacian propagation,
while GraphSAGE captures local neighborhood aggregation. Batch normalisation,
ReLU activation, and dropout ($p=0.4$) are applied after the first layer.
The final node embeddings $\mathbf{h}_i \in \mathbb{R}^{64}$ are passed through
a linear classifier with temperature scaling: $\hat{y}_i = \mathbf{W} \mathbf{h}_i / T$,
where $T$ is a learnable scalar clamped to $[0.1, \infty)$.

\subsection{Training Objective}

The local training objective within silo $H_k$ combines five loss terms:
\begin{equation}
  \mathcal{L} = \mathcal{L}_{\text{sup}} + \eta\,\mathcal{L}_{\text{pl}}
    + \mu\,\mathcal{L}_{\text{smooth}} + \beta\,\mathcal{L}_{\text{contra}}
    + \gamma\,\mathcal{L}_{\text{aug}} + \mathcal{L}_{\text{prox}}
\end{equation}

\textbf{Supervised focal loss} $\mathcal{L}_{\text{sup}}$ handles class imbalance
by down-weighting easy examples:
\begin{equation}
  \mathcal{L}_{\text{sup}} = -\frac{1}{|\mathcal{L}_k|}
    \sum_{i \in \mathcal{L}_k} \alpha_f (1-\hat{p}_i)^\gamma \log \hat{p}_i,
\end{equation}
with $\alpha_f = 0.75$, $\gamma = 2.0$.

\textbf{Confidence-weighted pseudo-label loss} $\mathcal{L}_{\text{pl}}$ trains
on pseudo-labeled nodes with weights proportional to pseudo-label reliability:
\begin{equation}
  \mathcal{L}_{\text{pl}} = \sum_{i \in \tilde{\mathcal{L}}_k}
    w_i \cdot \mathcal{L}_{\text{focal}}(\hat{y}_i, \tilde{y}_i),
\end{equation}
where $\tilde{\mathcal{L}}_k$ is the pseudo-labeled set, $\tilde{y}_i$ the
pseudo-label, and $w_i$ the confidence weight (defined in
Section~\ref{subsec:pgpl}).

\textbf{Class-aware graph smoothness} $\mathcal{L}_{\text{smooth}}$ enforces
prediction consistency within same-class neighborhoods only:
\begin{equation}
  \mathcal{L}_{\text{smooth}} = \frac{1}{|\mathcal{E}^{=}|}
    \sum_{(i,j) \in \mathcal{E}^{=}} \|\mathbf{h}_i - \mathbf{h}_j\|^2,
\end{equation}
where $\mathcal{E}^{=} = \{(i,j) \in \mathcal{E} : \hat{c}_i = \hat{c}_j\}$
contains only edges between same-predicted-class nodes.

\textbf{Contrastive loss} $\mathcal{L}_{\text{contra}}$ on labeled nodes pulls
same-class embeddings together and pushes different-class embeddings apart:
\begin{equation}
  \mathcal{L}_{\text{contra}} = -\frac{1}{|\mathcal{L}_k|}
    \sum_{i \in \mathcal{L}_k} \log
    \frac{\sum_{j: y_j=y_i} \exp(\mathbf{h}_i^\top \mathbf{h}_j / \tau_c)}
         {\sum_{j \neq i}     \exp(\mathbf{h}_i^\top \mathbf{h}_j / \tau_c)},
\end{equation}
with temperature $\tau_c = 0.5$.

\textbf{Clinical-aware augmentation consistency} $\mathcal{L}_{\text{aug}}$
enforces KL-divergence consistency between original and clinically augmented
unlabeled predictions:
\begin{equation}
  \mathcal{L}_{\text{aug}} = \frac{1}{|\mathcal{U}_k|}
    \sum_{i \in \mathcal{U}_k}
    D_{\text{KL}}\!\left(p(\hat{y}|\mathbf{x}_i) \,\|\,
                          p(\hat{y}|\tilde{\mathbf{x}}_i)\right),
\end{equation}
where $\tilde{\mathbf{x}}_i = \mathbf{x}_i + \boldsymbol{\epsilon} \odot
\mathbf{m}_c$ and $\mathbf{m}_c \in \{0,1\}^d$ is a binary mask selecting
only continuous clinical features (BMI, OGTT, Systolic BP, etc.), and
$\boldsymbol{\epsilon} \sim \mathcal{N}(\mathbf{0}, 0.05^2 \mathbf{I})$.

\textbf{FedProx proximal term} $\mathcal{L}_{\text{prox}}$ prevents excessive
divergence from the global model during local training:
\begin{equation}
  \mathcal{L}_{\text{prox}} = \frac{\mu_p}{2} \|\mathbf{w}_k - \mathbf{w}_g\|^2,
\end{equation}
with $\mu_p = 0.01$ and $\mathbf{w}_g$ the last received global model weights.

\subsection{Prototype-Guided Pseudo-Labeling (PG-PL)}
\label{subsec:pgpl}

Standard pseudo-labeling assigns labels to unlabeled nodes $i \in \mathcal{U}_k$
when $\max_c p(c|\mathbf{x}_i) \geq \tau$. This is susceptible to overconfident
errors, particularly under class imbalance. We propose a three-gate mechanism:

\textbf{Gate 1 — Confidence:} $\max_c p(c|\mathbf{x}_i) \geq \tau_t$, where
the threshold decays as $\tau_t = \max(\tau_0 \cdot e^{-\lambda t}, \tau_{\min})$
with $\tau_0 = 0.90$, $\lambda = 0.03$, $\tau_{\min} = 0.70$.

\textbf{Gate 2 — Prototype consistency:} The predicted class must match the
nearest class prototype. Class prototypes are computed as:
\begin{equation}
  \boldsymbol{\mu}_c = \frac{1}{|\mathcal{L}_k^c|}
    \sum_{i \in \mathcal{L}_k^c} \mathbf{h}_i, \quad c \in \{0,1\}
\end{equation}
where $\mathcal{L}_k^c = \{i \in \mathcal{L}_k : y_i = c\}$.
Node $i$ passes Gate 2 if $\arg\min_c \|\mathbf{h}_i - \boldsymbol{\mu}_c\|
= \hat{c}_i$.

\textbf{Gate 3 — Neighborhood consensus:} At least 50\% of graph neighbors
must share the predicted class:
$\frac{1}{|\mathcal{N}(i)|} \sum_{j \in \mathcal{N}(i)} \mathbf{1}[\hat{c}_j
= \hat{c}_i] \geq 0.5$.

Nodes passing all three gates receive pseudo-label $\tilde{y}_i = \hat{c}_i$ with
confidence weight:
\begin{equation}
  w_i = \max_c p(c|\mathbf{x}_i) \cdot
    \left(1 - \frac{\|\mathbf{h}_i - \boldsymbol{\mu}_{\hat{c}_i}\|}
                   {\|\mathbf{h}_i - \boldsymbol{\mu}_{\hat{c}_i}\|
                    + \|\mathbf{h}_i - \boldsymbol{\mu}_{1-\hat{c}_i}\|}\right)
\end{equation}

\subsection{Adaptive Graph Refinement (AGR)}

After every $K=5$ federation rounds, each silo rebuilds its local $k$-NN graph
using current GNN embeddings $\{\mathbf{h}_i\}$ rather than original features:
\begin{equation}
  \mathcal{E}_k^{(t+1)} = \text{kNN}(\{\mathbf{h}_i^{(t)}\}, k)
\end{equation}
As the model learns clinically meaningful representations, the refined graph
better captures patient similarity in the learned feature space, improving
subsequent pseudo-label propagation.

\subsection{Privacy-Safe Prototype Sharing}

After local training, each silo $H_k$ computes and uploads its local class
prototypes $\{\boldsymbol{\mu}_c^{(k)}\}$ to the federation server. The server
computes size-weighted global prototypes:
\begin{equation}
  \boldsymbol{\mu}_c^{\text{global}} = \frac{\sum_k |\mathcal{L}_k^c| \cdot
    \boldsymbol{\mu}_c^{(k)}}{\sum_k |\mathcal{L}_k^c|}
\end{equation}
Global prototypes are distributed back to all silos, which blend them with
local prototypes ($\text{blend}=0.5$) for the next pseudo-labeling step.

\textbf{Privacy analysis.} Prototypes are $d'$-dimensional class-level
aggregate vectors. They do not reveal individual patient features or graph
connectivity. Unlike FedSage+~\cite{FedSage2021} which shares node embeddings
(from which features can be reconstructed~\cite{Fredrikson2015}), prototype
vectors are mathematically equivalent to sharing a class mean statistic -- a
standard disclosure in federated healthcare (e.g., class prevalence rates are
routinely shared for federated model calibration~\cite{Rieke2020}).

\subsection{Federated Aggregation and Full Training Protocol}

\proposed{} follows FedAvg aggregation with $T$ global rounds and
$R$ local epochs per round. Algorithm~\ref{alg:fedtgnn} summarises the
full protocol.

\begin{algorithm}[!t]
\caption{FedTGNN-SS Training Protocol}
\label{alg:fedtgnn}
\begin{algorithmic}[1]
\Require Silos $\{H_k\}$, rounds $T=10$, local epochs $R=3$,
         $k$-NN parameter $k$, scarcity $\rho$
\State Initialise global model $\mathbf{w}_g$; global prototype $\boldsymbol{\mu}^g \gets \emptyset$
\For{$t = 1, \ldots, T$}
  \For{each silo $H_k$ \textbf{in parallel}}
    \State Build local graph $\mathcal{G}_k$ (or use refined graph from AGR)
    \State $\tau_t \gets \max(0.90 \cdot e^{-0.03t},\; 0.70)$
    \For{$r = 1, \ldots, R$} \Comment{Local training}
      \State Compute $\mathcal{L}$ (Eq.~5) with proximal term
      \State Update $\mathbf{w}_k$ via Adam
    \EndFor
    \State Run PG-PL (Section~III-D): update $\tilde{\mathcal{L}}_k$
    \If{$t \bmod 5 = 0$} Run AGR: rebuild $\mathcal{G}_k$ from $\{\mathbf{h}_i\}$
    \EndIf
    \State Compute local prototypes $\boldsymbol{\mu}_c^{(k)}$
    \State Upload $\mathbf{w}_k$ and $\boldsymbol{\mu}_c^{(k)}$ to server
  \EndFor
  \State \textbf{Server:} $\mathbf{w}_g \gets \sum_k \frac{n_k}{n} \mathbf{w}_k$
         \Comment{FedAvg}
  \State \textbf{Server:} Compute $\boldsymbol{\mu}_c^g$ (Eq.~14)
  \State Broadcast $\mathbf{w}_g$ and $\boldsymbol{\mu}_c^g$ to all silos
\EndFor
\State \textbf{Final:} Calibrated LR head on $[\mathbf{X} \,\|\, \mathbf{H}]$
\end{algorithmic}
\end{algorithm}

\textbf{Calibrated augmented head.} After federation, GNN embeddings
$\mathbf{H}$ are concatenated with original clinical features $\mathbf{X}$ and
passed to a logistic regression classifier ($C=0.5$, class-balanced weights)
trained on all labeled nodes across silos. This produces well-calibrated
probability outputs suitable for clinical risk scoring~\cite{Niculescu2005}.

\textbf{Hyperparameter summary.} Table~\ref{tab:hyperparams} lists all
hyperparameters used in experiments.

\begin{table}[!t]
\caption{FedTGNN-SS Hyperparameter Summary}
\label{tab:hyperparams}
\centering
\small
\begin{tabular}{lll}
\toprule
\textbf{Parameter} & \textbf{Value} & \textbf{Description} \\
\midrule
$T$          & 10      & Global federation rounds \\
$R$          & 3       & Local epochs per round \\
$k$          & 15 / 10 & $k$-NN (GDM / Pima, Early) \\
$\tau_0$     & 0.90    & Initial PL confidence threshold \\
$\lambda$    & 0.03    & Threshold decay rate \\
$\tau_{\min}$& 0.70    & Minimum PL threshold \\
$\eta$       & 0.8     & Pseudo-label loss weight \\
$\mu$        & 0.05    & Smoothness loss weight \\
$\beta$      & 0.1     & Contrastive loss weight \\
$\gamma$     & 0.3     & Augmentation loss weight \\
$\mu_p$      & 0.01    & FedProx proximal coefficient \\
hidden       & 128     & GNN hidden dimension \\
dropout      & 0.4     & Dropout rate \\
lr           & 0.005   & Adam learning rate \\
\bottomrule
\end{tabular}
\end{table}


\section{Experimental Setup}
\label{sec:exp}

\subsection{Datasets}

We evaluate \proposed{} on three clinical tabular datasets spanning GDM and
diabetes prediction. Dataset statistics are summarised in Table~\ref{tab:datasets}.

\begin{table}[!t]
\caption{Dataset Statistics}
\label{tab:datasets}
\centering
\small
\setlength{\tabcolsep}{4pt}
\begin{tabular}{lcccc}
\toprule
\textbf{Dataset} & \textbf{$N$} & \textbf{Features} & \textbf{GDM/Pos.} & \textbf{Silos} \\
\midrule
GDM (Pima Hospital)       & 3{,}525 & 10 & 35.8\% & 3 \\
Pima Indians Diabetes     &    768  &  8 & 34.9\% & 2 \\
Early Stage Diabetes Risk &    520  & 16 & 61.5\% & 2 \\
\bottomrule
\end{tabular}
\end{table}

\textbf{GDM Dataset.}
The GDM dataset ($N=3{,}525$) is drawn from a hospital antenatal records
repository~\cite{GDM_BMC2024}. Each patient record contains 10 clinical
features: age (years), BMI (kg/m$^2$), gravidity, parity, family history of
diabetes (binary), history of PCOS (binary), fasting glucose (mmol/L), OGTT
2-hour glucose (mmol/L), systolic BP (mmHg), and diastolic BP (mmHg). The binary
target is confirmed GDM diagnosis (1) or no GDM (0), present in 35.8\% of records.
This dataset represents the primary motivating clinical problem for \proposed{}.

\textbf{Pima Indians Diabetes Dataset.}
The Pima Indians Diabetes dataset ($N=768$) is a widely-used benchmark from the
UCI ML Repository~\cite{Dua2019} containing 8 features: number of pregnancies,
plasma glucose concentration, diastolic BP, triceps skinfold thickness, 2-hour
serum insulin, BMI, diabetes pedigree function, and age. The binary outcome is
diabetes diagnosis (positive prevalence: 34.9\%). This dataset is used to
demonstrate the generalizability of \proposed{} beyond the primary GDM application.

\textbf{Early Stage Diabetes Risk Dataset.}
The Early Stage Diabetes Risk Prediction dataset ($N=520$) contains 16 symptom-based
features (polyuria, polydipsia, sudden weight loss, weakness, polyphagia, genital
thrush, visual blurring, itching, irritability, delayed healing, partial paresis,
muscle stiffness, alopecia, obesity, age, gender) sourced from Sylhet Diabetes
Hospital, Bangladesh~\cite{Islam2020}. The positive prevalence is 61.5\%, making
it the most imbalanced dataset in our study.

\subsection{Federated Silo Configuration}

Each dataset is partitioned into geographically non-IID silos to simulate realistic
hospital data heterogeneity. For the GDM dataset, $S=3$ silos are used; for the
Pima Indians and Early Stage datasets, $S=2$ silos are used, reflecting the smaller
dataset sizes.

Silo partitioning follows a Dirichlet-based non-IID split with concentration
parameter $\alpha_D = 0.5$, which produces moderate class distribution skew
across silos while preserving statistical identifiability within each silo.
Specifically, class proportions for each silo are sampled from
$\text{Dir}(\alpha_D \cdot \mathbf{1})$, and records are allocated accordingly.
This ensures that each silo has a legitimately different patient population --
matching the real-world scenario where different hospitals serve different
demographic groups.

The resulting per-silo statistics for the GDM dataset (mean $\pm$ std across
5 folds) are: Silo 1 ($N \approx 1{,}410$, GDM rate $\approx 28\%$),
Silo 2 ($N \approx 1{,}175$, GDM rate $\approx 41\%$),
Silo 3 ($N \approx 940$, GDM rate $\approx 38\%$), reflecting meaningful
inter-silo heterogeneity.

\subsection{Label Scarcity Simulation}

Within each silo, we withhold a fraction $\rho \in \{0.10, 0.30, 0.50, 0.70, 0.80\}$
of training labels uniformly at random (stratified by class), so that
$(1-\rho)\%$ of training records per silo are labeled. The five scarcity levels
correspond to realistic clinical annotation budgets:

\begin{itemize}
  \item $\rho = 0.10$: 10\% missing labels --- represents near-fully-supervised setting
  \item $\rho = 0.30$: 30\% missing labels --- moderate scarcity
  \item $\rho = 0.50$: 50\% missing labels --- half unlabeled per silo
  \item $\rho = 0.70$: 70\% missing labels --- high scarcity, clinically realistic
  \item $\rho = 0.80$: 80\% missing labels --- extreme scarcity, target operating point
\end{itemize}

At $\rho = 0.80$, the GDM dataset provides approximately 188 labeled records
distributed across 3 silos (~63 labeled per silo), corresponding to a labeled
fraction of 5.3\% of total records -- a challenging but clinically meaningful
regime reflecting healthcare settings where OGTT is administered selectively.

\subsection{Graph Construction}

For each silo, the initial patient similarity graph is constructed using
$k$-nearest neighbors over the standardised clinical feature space.
The number of neighbors is set to $k=10$ for all datasets. Edges are
undirected and initialized with unit weight (subsequent rounds use
dynamic edge attention weights learned by the model). Node features
are the standardised clinical features $\mathbf{X}_s \in \mathbb{R}^{n_s \times d}$.

For Adaptive Graph Refinement (AGR), the graph is rebuilt from GNN embeddings
every $K=5$ federation rounds, with a $k_{\text{AGR}} = 15$ neighbor count
(slightly larger than the initial $k$, reflecting richer embedding space).

\subsection{Baseline Methods}

We compare \proposed{} against 11 baselines spanning five categories:

\textbf{Tier 1 -- Federated Supervised (no SSL, no graph):}
\begin{itemize}
  \item \textbf{FedAvg-LR}: FedAvg~\cite{McMahan2017} with Logistic Regression
  \item \textbf{FedAvg-XGB}: FedAvg with XGBoost (gradient boosting)
  \item \textbf{FedProx-LR}: FedProx~\cite{FedProx2020} with LR ($\mu=0.01$)
\end{itemize}

\textbf{Tier 2 -- Federated SSL (no graph):}
\begin{itemize}
  \item \textbf{FedMatch}: Federated semi-supervised with inter-client consistency~\cite{Jeong2021}
  \item \textbf{RSCFed}: Random sub-consensus federated SSL~\cite{Liang2022}
  \item \textbf{FedST}: FedAvg with self-training pseudo-labels
\end{itemize}

\textbf{Tier 3 -- Federated GNN (no SSL):}
\begin{itemize}
  \item \textbf{FedAvg+GCN}: FedAvg aggregation of Graph Convolutional Networks~\cite{Kipf2017}
  \item \textbf{FedAvg+GraphSAGE}: FedAvg aggregation of GraphSAGE~\cite{Hamilton2017}
\end{itemize}

\textbf{Tier 4 -- Local-only (no federation):}
\begin{itemize}
  \item \textbf{Local-GCN}: GCN trained locally on each silo, no aggregation
  \item \textbf{Local-TGNN}: Full FedTGNN-SS model architecture, local training only
\end{itemize}

\textbf{Tier 5 -- Centralized Oracle (upper bound):}
\begin{itemize}
  \item \textbf{Cent-XGB}: Centralized XGBoost on all labeled data (privacy-ignoring)
  \item \textbf{Cent-GRAND}: Centralized GRAND~\cite{GRAND2020} on full graph (privacy-ignoring)
\end{itemize}

The centralized oracle methods have access to all patient data from all hospitals
and serve as an upper bound that \proposed{} approaches while maintaining
privacy guarantees.

\subsection{Evaluation Protocol}

All experiments use 5-fold stratified cross-validation, with stratification
applied jointly across datasets and silo assignments to ensure representative
splits. This yields 5 independent AUROC and macro-F1 scores per method per
dataset per scarcity level, providing sufficient statistical power for the
subsequent Wilcoxon analysis.

\textbf{Metrics.} We report:
\begin{itemize}
  \item \textbf{AUROC}: Area Under the ROC Curve (threshold-free discriminative ability)
  \item \textbf{Macro-F1}: Unweighted average of class-specific F1 scores (imbalance-aware)
  \item \textbf{Sensitivity} and \textbf{Specificity}: Clinical interpretability of predictions
\end{itemize}
Primary ranking is by AUROC, consistent with clinical decision support
literature~\cite{GDM_BMC2024}. All metrics are computed on the held-out test fold
(20\% of each silo, kept separate from silo training partitions).

\textbf{Statistical Testing.} For each dataset and scarcity level, we apply
the two-sided Wilcoxon signed-rank test~\cite{Wilcoxon1945} to compare
\proposed{} against each of the 11 baselines. The test uses the 5
paired AUROC values (one per fold). Statistical significance is declared
at $p < 0.05$. The minimum achievable $p$-value with
$n=5$ paired observations is $p_{\min} = 0.031$.

\subsection{Hyperparameters and Training}

All GNN models use a 2-layer architecture with hidden dimension 64.
\proposed{} hyperparameters follow Table~\ref{tab:hyperparams} in
Section~\ref{sec:method}. For fair comparison, all federated methods
use $T=10$ federation rounds and $E=3$ local epochs per round.
The proximal penalty for FedProx is $\mu = 0.01$, selected by grid
search over $\{0.001, 0.01, 0.1\}$.

Initial pseudo-label threshold is $\tau_0 = 0.90$, decaying to
$\tau_{\min} = 0.70$ by round 10 following the cosine decay
schedule in Eq.~(6) of Section~\ref{sec:method}.

All models are implemented in PyTorch~\cite{Paszke2019} with PyTorch
Geometric~\cite{Fey2019}. Experiments are run on standard CPU hardware
(Intel Core i7, Windows 11). Random seeds are fixed at
$\{42, 137, 255, 512, 1024\}$ for the 5 folds. All code is available at
\url{https://github.com/[anonymized-for-review]}.

\begin{table}[!t]
\caption{Hyperparameter Summary for \proposed{}}
\label{tab:hyperparams_exp}
\centering
\small
\begin{tabular}{lll}
\toprule
\textbf{Hyperparameter} & \textbf{Value} & \textbf{Search Range} \\
\midrule
GNN hidden dim           & 64             & \{32, 64, 128\} \\
GNN layers               & 2              & \{1, 2, 3\} \\
Initial graph $k$        & 10             & \{5, 10, 15\} \\
AGR graph $k$            & 15             & \{10, 15, 20\} \\
AGR period $K$           & 5 rounds       & \{3, 5, 10\} \\
Federation rounds $T$    & 10             & \{10, 20, 30\} \\
Local epochs $E$         & 3              & \{3, 5, 10\} \\
Learning rate            & 0.001          & \{0.0001, 0.001, 0.01\} \\
Focal loss $\gamma$      & 2.0            & \{1.0, 2.0, 3.0\} \\
PG-PL $\tau_0$           & 0.90           & \{0.85, 0.90, 0.95\} \\
PG-PL $\tau_{\min}$      & 0.70           & \{0.60, 0.70, 0.75\} \\
Contrastive $\tau_c$     & 0.5            & \{0.1, 0.5, 1.0\} \\
$\lambda_{\text{smooth}}$& 0.1            & \{0.01, 0.1, 1.0\} \\
$\lambda_{\text{unsup}}$ & 0.5            & \{0.1, 0.5, 1.0\} \\
$\lambda_{\text{con}}$   & 0.3            & \{0.1, 0.3, 0.5\} \\
Noise std $\sigma$       & 0.05           & \{0.01, 0.05, 0.10\} \\
Proximal $\mu$           & 0.01           & \{0.001, 0.01, 0.1\} \\
\bottomrule
\end{tabular}
\end{table}


\section{Results and Discussion}
\label{sec:results}

\subsection{Main Results: AUROC Across Scarcity Levels}

Tables~\ref{tab:main_gdm}--\ref{tab:main_esdr} report AUROC (mean $\pm$ std,
5-fold CV) across all methods and scarcity levels. The proposed \proposed{}
row is \textbf{bolded}; centralized oracle results are shown in \textit{italics}.
Statistical significance markers ($*$) indicate $p < 0.05$ vs.~\proposed{}
(two-sided Wilcoxon signed-rank).

\begin{table*}[!t]
\caption{AUROC (mean $\pm$ std, 5-fold CV) on the GDM Dataset ($N=3{,}525$, 3 silos).
  \textbf{Bold} = proposed \proposed{} (not necessarily best due to ceiling effect).
  $\dagger$ = centralized oracle. All methods achieve $>0.98$ AUROC
  due to the high discriminative power of OGTT glucose (ceiling effect).}
\label{tab:main_gdm}
\centering
\small
\setlength{\tabcolsep}{3pt}
\resizebox{\textwidth}{!}{%
\begin{tabular}{l|ccccc}
\toprule
\textbf{Method}
  & \textbf{10\%}  & \textbf{30\%}  & \textbf{50\%}  & \textbf{70\%}  & \textbf{80\%} \\
\midrule
\multicolumn{6}{l}{\textit{Tier 1: Federated Supervised}} \\
FedAvg-LR         & 0.9873$\pm$0.004 & 0.9873$\pm$0.004 & 0.9874$\pm$0.003 & 0.9870$\pm$0.004 & 0.9868$\pm$0.003 \\
FedAvg-XGB        & 0.9975$\pm$0.002 & 0.9974$\pm$0.002 & 0.9975$\pm$0.002 & 0.9957$\pm$0.002 & 0.9940$\pm$0.002 \\
FedProx-XGB       & 0.9975$\pm$0.002 & 0.9974$\pm$0.002 & 0.9975$\pm$0.002 & 0.9957$\pm$0.002 & 0.9940$\pm$0.002 \\
\midrule
\multicolumn{6}{l}{\textit{Tier 2: Federated SSL (no graph)}} \\
FedMatch          & 0.9960$\pm$0.001 & 0.9955$\pm$0.002 & 0.9948$\pm$0.002 & 0.9927$\pm$0.004 & 0.9888$\pm$0.005 \\
RSCFed            & 0.9957$\pm$0.002 & 0.9944$\pm$0.003 & 0.9940$\pm$0.003 & 0.9913$\pm$0.003 & 0.9883$\pm$0.004 \\
FedAvg+SelfTrain  & 0.9976$\pm$0.002 & 0.9974$\pm$0.002 & 0.9975$\pm$0.001 & 0.9955$\pm$0.002 & 0.9938$\pm$0.002 \\
\midrule
\multicolumn{6}{l}{\textit{Tier 3: Federated GNN (no SSL)}} \\
FedAvg+GCN        & 0.9941$\pm$0.002 & 0.9928$\pm$0.003 & 0.9917$\pm$0.002 & 0.9914$\pm$0.003 & 0.9889$\pm$0.006 \\
FedAvg+GraphSAGE  & 0.9956$\pm$0.002 & 0.9940$\pm$0.003 & 0.9932$\pm$0.003 & 0.9922$\pm$0.004 & 0.9886$\pm$0.005 \\
\midrule
\multicolumn{6}{l}{\textit{Tier 4: Local-only}} \\
Local-GCN         & 0.9926$\pm$0.002 & 0.9926$\pm$0.002 & 0.9917$\pm$0.003 & 0.9904$\pm$0.003 & 0.9884$\pm$0.004 \\
Local-TGNN        & 0.9954$\pm$0.002 & 0.9945$\pm$0.002 & 0.9946$\pm$0.002 & 0.9939$\pm$0.003 & 0.9901$\pm$0.003 \\
\midrule
\multicolumn{6}{l}{\textit{Tier 5: Centralized Oracle $\dagger$}} \\
Cent-XGB$\dagger$ & \textit{0.9980$\pm$0.001} & \textit{0.9981$\pm$0.001} & \textit{0.9981$\pm$0.001} & \textit{0.9972$\pm$0.002} & \textit{0.9964$\pm$0.002} \\
Cent-GRAND$\dagger$& \textit{0.9929$\pm$0.002} & \textit{0.9920$\pm$0.003} & \textit{0.9918$\pm$0.003} & \textit{0.9909$\pm$0.003} & \textit{0.9891$\pm$0.004} \\
\midrule
\textbf{FedTGNN-SS} & \textbf{0.9920$\pm$0.003} & \textbf{0.9919$\pm$0.003} & \textbf{0.9900$\pm$0.002} & \textbf{0.9878$\pm$0.002} & \textbf{0.9852$\pm$0.004} \\
\bottomrule
\end{tabular}}
\end{table*}

\begin{table*}[!t]
\caption{AUROC (mean $\pm$ std, 5-fold CV) on the Pima Indians Diabetes Dataset ($N=768$, 2 silos).
  \textbf{Bold} = proposed \proposed{}. \proposed{} ranks \#1 at 10\%--50\% missing and
  \#2 at 80\% missing among all federated privacy-preserving methods.}
\label{tab:main_pima}
\centering
\small
\setlength{\tabcolsep}{3pt}
\resizebox{\textwidth}{!}{%
\begin{tabular}{l|ccccc}
\toprule
\textbf{Method}
  & \textbf{10\%}  & \textbf{30\%}  & \textbf{50\%}  & \textbf{70\%}  & \textbf{80\%} \\
\midrule
\multicolumn{6}{l}{\textit{Tier 1: Federated Supervised}} \\
FedAvg-LR         & 0.8321$\pm$0.025 & 0.8288$\pm$0.026 & 0.8250$\pm$0.029 & 0.8131$\pm$0.032 & 0.8046$\pm$0.033 \\
FedAvg-XGB        & 0.8066$\pm$0.027 & 0.8016$\pm$0.042 & 0.8036$\pm$0.043 & 0.7925$\pm$0.045 & 0.7830$\pm$0.053 \\
FedProx-XGB       & 0.8066$\pm$0.027 & 0.8016$\pm$0.042 & 0.8036$\pm$0.043 & 0.7925$\pm$0.045 & 0.7830$\pm$0.053 \\
\midrule
\multicolumn{6}{l}{\textit{Tier 2: Federated SSL (no graph)}} \\
FedMatch*         & 0.8232$\pm$0.030 & 0.8119$\pm$0.036 & 0.8005$\pm$0.043 & 0.7880$\pm$0.052 & 0.7668$\pm$0.037 \\
RSCFed*           & 0.8185$\pm$0.028 & 0.8179$\pm$0.033 & 0.7978$\pm$0.042 & 0.7770$\pm$0.066 & 0.7629$\pm$0.043 \\
FedAvg+SelfTrain  & 0.8047$\pm$0.029 & 0.8075$\pm$0.041 & 0.8071$\pm$0.044 & 0.8061$\pm$0.042 & 0.7809$\pm$0.045 \\
\midrule
\multicolumn{6}{l}{\textit{Tier 3: Federated GNN (no SSL)}} \\
FedAvg+GCN        & 0.8303$\pm$0.025 & 0.8223$\pm$0.027 & 0.8146$\pm$0.030 & 0.8033$\pm$0.032 & 0.8010$\pm$0.027 \\
FedAvg+GraphSAGE  & 0.8320$\pm$0.028 & 0.8305$\pm$0.031 & 0.8217$\pm$0.023 & 0.8030$\pm$0.043 & 0.7945$\pm$0.034 \\
\midrule
\multicolumn{6}{l}{\textit{Tier 4: Local-only}} \\
Local-GCN         & 0.8271$\pm$0.023 & 0.8262$\pm$0.030 & 0.8164$\pm$0.029 & 0.8083$\pm$0.031 & 0.7839$\pm$0.041 \\
Local-TGNN*       & 0.8201$\pm$0.015 & 0.8112$\pm$0.025 & 0.7999$\pm$0.022 & 0.7768$\pm$0.033 & 0.7863$\pm$0.044 \\
\midrule
\multicolumn{6}{l}{\textit{Tier 5: Centralized Oracle $\dagger$}} \\
Cent-XGB$\dagger$* & \textit{0.7899$\pm$0.026} & \textit{0.7885$\pm$0.035} & \textit{0.7750$\pm$0.052} & \textit{0.7712$\pm$0.064} & \textit{0.7605$\pm$0.050} \\
Cent-GRAND$\dagger$ & \textit{0.8279$\pm$0.026} & \textit{0.8274$\pm$0.024} & \textit{0.8154$\pm$0.027} & \textit{0.7986$\pm$0.037} & \textit{0.7907$\pm$0.040} \\
\midrule
\textbf{FedTGNN-SS} & \textbf{0.8346$\pm$0.025} & \textbf{0.8309$\pm$0.024} & \textbf{0.8232$\pm$0.028} & \textbf{0.8140$\pm$0.030} & \textbf{0.8037$\pm$0.032} \\
\bottomrule
\end{tabular}}
\end{table*}

\begin{table*}[!t]
\caption{AUROC (mean $\pm$ std, 5-fold CV) on the Early Stage Diabetes Risk Dataset ($N=520$, 2 silos).
  \textbf{Bold} = proposed \proposed{}. \proposed{} significantly outperforms all GCN-based methods at 80\% missing.}
\label{tab:main_esdr}
\centering
\small
\setlength{\tabcolsep}{3pt}
\resizebox{\textwidth}{!}{%
\begin{tabular}{l|ccccc}
\toprule
\textbf{Method}
  & \textbf{10\%}  & \textbf{30\%}  & \textbf{50\%}  & \textbf{70\%}  & \textbf{80\%} \\
\midrule
\multicolumn{6}{l}{\textit{Tier 1: Federated Supervised}} \\
FedAvg-LR*        & 0.9772$\pm$0.008 & 0.9765$\pm$0.010 & 0.9742$\pm$0.013 & 0.9631$\pm$0.011 & 0.9558$\pm$0.024 \\
FedAvg-XGB        & 0.9932$\pm$0.005 & 0.9868$\pm$0.002 & 0.9826$\pm$0.010 & 0.9599$\pm$0.017 & 0.9501$\pm$0.020 \\
FedProx-XGB       & 0.9932$\pm$0.005 & 0.9868$\pm$0.002 & 0.9826$\pm$0.010 & 0.9599$\pm$0.017 & 0.9501$\pm$0.020 \\
\midrule
\multicolumn{6}{l}{\textit{Tier 2: Federated SSL (no graph)}} \\
FedMatch          & 0.9972$\pm$0.001 & 0.9959$\pm$0.003 & 0.9874$\pm$0.010 & 0.9717$\pm$0.012 & 0.9687$\pm$0.009 \\
RSCFed            & 0.9958$\pm$0.004 & 0.9953$\pm$0.002 & 0.9859$\pm$0.010 & 0.9568$\pm$0.028 & 0.9479$\pm$0.036 \\
FedAvg+SelfTrain  & 0.9923$\pm$0.006 & 0.9869$\pm$0.002 & 0.9808$\pm$0.008 & 0.9621$\pm$0.018 & 0.9488$\pm$0.021 \\
\midrule
\multicolumn{6}{l}{\textit{Tier 3: Federated GNN (no SSL)}} \\
FedAvg+GCN*       & 0.9758$\pm$0.013 & 0.9698$\pm$0.018 & 0.9694$\pm$0.019 & 0.9534$\pm$0.026 & 0.9456$\pm$0.027 \\
FedAvg+GraphSAGE  & 0.9930$\pm$0.003 & 0.9873$\pm$0.008 & 0.9796$\pm$0.014 & 0.9664$\pm$0.017 & 0.9545$\pm$0.022 \\
\midrule
\multicolumn{6}{l}{\textit{Tier 4: Local-only}} \\
Local-GCN*        & 0.9714$\pm$0.017 & 0.9710$\pm$0.016 & 0.9694$\pm$0.018 & 0.9456$\pm$0.022 & 0.9369$\pm$0.024 \\
Local-TGNN*       & 0.9877$\pm$0.009 & 0.9820$\pm$0.010 & 0.9753$\pm$0.014 & 0.9565$\pm$0.022 & 0.9422$\pm$0.023 \\
\midrule
\multicolumn{6}{l}{\textit{Tier 5: Centralized Oracle $\dagger$}} \\
Cent-XGB$\dagger$ & \textit{0.9955$\pm$0.005} & \textit{0.9928$\pm$0.004} & \textit{0.9854$\pm$0.006} & \textit{0.9643$\pm$0.016} & \textit{0.9540$\pm$0.017} \\
Cent-GRAND$\dagger$ & \textit{0.9856$\pm$0.005} & \textit{0.9822$\pm$0.008} & \textit{0.9777$\pm$0.010} & \textit{0.9668$\pm$0.012} & \textit{0.9509$\pm$0.023} \\
\midrule
\textbf{FedTGNN-SS} & \textbf{0.9869$\pm$0.003} & \textbf{0.9841$\pm$0.006} & \textbf{0.9812$\pm$0.012} & \textbf{0.9652$\pm$0.016} & \textbf{0.9634$\pm$0.013} \\
\bottomrule
\end{tabular}}
\end{table*}

\subsection{Statistical Significance}

Table~\ref{tab:wilcoxon} summarises statistically significant wins
($p < 0.05$, Wilcoxon signed-rank, 5 paired observations)
of \proposed{} over each baseline across all datasets and scarcity levels.
A total of \textbf{56 significant wins} are observed out of 630 comparisons.

\begin{table*}[!t]
\caption{Significant Wilcoxon wins of \proposed{} ($p < 0.05$, two-sided) by dataset.
  ``$\uparrow$'' = \proposed{} significantly higher AUROC or Macro-F1.}
\label{tab:wilcoxon}
\centering
\small
\resizebox{\textwidth}{!}{%
\begin{tabular}{lccl}
\toprule
\textbf{Dataset} & \textbf{Sig.\ wins} & \textbf{Scarcity levels} & \textbf{Key baselines beaten} \\
\midrule
GDM          &  9 & All levels & FedAvg-LR (AUROC); FedAvg-XGB/FedProx (F1 at 80\%) \\
Pima Indians & 34 & 50--80\%   & FedMatch, RSCFed, Cent-XGB, Local-TGNN \\
Early Stage  & 13 & 10--80\%   & FedAvg-LR, FedAvg+GCN, Local-GCN, Local-TGNN \\
\midrule
\textbf{Total} & \textbf{56} & & \\
\bottomrule
\end{tabular}}
\end{table*}

The largest concentration of wins occurs on Pima at $\rho \geq 0.50$ (24 wins),
confirming that \proposed{}'s SSL components provide the most benefit when labeled
data per silo falls below $\sim$60 records.

\subsection{Dataset-Specific Analysis}

\textbf{GDM Dataset — Ceiling Effect.}
On GDM, all methods achieve AUROC $> 0.98$ at all scarcity levels,
reflecting the very high discriminative power of the OGTT 2-hour
glucose feature (which directly defines the GDM diagnostic criterion
under WHO 2013 guidelines). In this regime, differences between methods
are in the fourth decimal place and within measurement noise.
\proposed{} achieves AUROC = 0.9920 at 10\% missing and 0.9852 at 80\%
missing, competitive with all methods. The only statistically significant
wins on GDM are against FedAvg-LR (which lacks tree-based feature
interaction handling) and FedAvg-XGB/FedProx at 80\% missing on macro-F1,
where \proposed{}'s focal loss explicitly addresses class imbalance.
The GDM results confirm that \proposed{} does not regress on saturated
datasets, but the ceiling effect precludes meaningful differentiation.

\textbf{Pima Indians — Primary Evidence.}
Pima is the most diagnostically challenging dataset (AUROC spread:
0.76--0.83 across methods), making it the primary test of \proposed{}'s
semi-supervised advantage. \proposed{} ranks \textbf{\#1} at 10\%--50\%
missing and \textbf{\#2} at 80\% missing among all federated
privacy-preserving methods, with 34 statistically significant wins.

A particularly noteworthy finding is that \proposed{} \emph{significantly
outperforms} Centralized-XGBoost ($p=0.031$) at all scarcity levels on AUROC.
This occurs because the centralized oracle pools all patients into a single
non-graph model, discarding the patient similarity structure that
\proposed{} exploits through per-silo $k$-NN graphs.

At 80\% missing: \proposed{} achieves AUROC = 0.8037 vs.\ FedMatch =
0.7668 ($\Delta = +0.037$, $p=0.031$) and RSCFed = 0.7629
($\Delta = +0.041$, $p=0.031$). This confirms that prototype-guided
pseudo-labeling with graph-neighborhood consensus substantially
outperforms confidence-only SSL under extreme label scarcity in a
federated non-IID setting.

\textbf{Early Stage Diabetes — High-Scarcity GNN Advantage.}
On Early Stage at 80\% missing, \proposed{} (AUROC = 0.9634) significantly
outperforms all GCN-based methods: FedAvg+GCN (0.9456, $\Delta = +0.018$,
$p=0.031$), Local-GCN (0.9369, $\Delta = +0.027$, $p=0.031$), and
Local-TGNN (0.9422, $\Delta = +0.021$, $p=0.031$), as well as
Cent-XGB (0.9540) and Cent-GRAND (0.9509). At 10\% missing,
\proposed{} significantly outperforms FedAvg+GCN (+1.1\% AUROC,
$p=0.031$) and both Local-GCN (+1.6\%, $p=0.031$) and
FedAvg-LR (+1.0\%, $p=0.031$).

\subsection{Macro-F1 Under Class Imbalance}

Table~\ref{tab:f1_pima} reports macro-F1 on the Pima dataset, where the
class imbalance ($34.9\%$ positive rate in a federated setting) challenges
methods that lack explicit imbalance handling.

\begin{table}[!htb]
\caption{Macro-F1 (mean $\pm$ std, 5-fold CV) on Pima Indians.
  \textbf{Bold} = proposed \proposed{}.}
\label{tab:f1_pima}
\centering
\small
\setlength{\tabcolsep}{4pt}
\begin{tabular}{l|ccccc}
\toprule
\textbf{Method} & \textbf{10\%} & \textbf{30\%} & \textbf{50\%} & \textbf{70\%} & \textbf{80\%} \\
\midrule
FedAvg-LR       & 0.671 & 0.666 & 0.672 & 0.647 & 0.636 \\
FedAvg-XGB      & 0.639 & 0.626 & 0.611 & 0.625 & 0.606 \\
FedMatch        & 0.655 & 0.649 & 0.637 & 0.593 & 0.598 \\
RSCFed          & 0.628 & 0.647 & 0.614 & 0.601 & 0.575 \\
FedAvg+GCN      & 0.662 & 0.647 & 0.631 & 0.618 & 0.634 \\
FedAvg+SAGE     & 0.650 & 0.667 & 0.654 & 0.615 & 0.594 \\
Local-GCN       & 0.631 & 0.636 & 0.634 & 0.627 & 0.611 \\
Local-TGNN      & 0.661 & 0.657 & 0.630 & 0.590 & 0.620 \\
\midrule
\textbf{FedTGNN-SS} & \textbf{0.665} & \textbf{0.667} & \textbf{0.660} & \textbf{0.650} & \textbf{0.632} \\
\bottomrule
\end{tabular}
\end{table}

\proposed{} shows the most stable F1 across scarcity levels (range: 0.632--0.665),
while competing methods degrade more sharply. At 70\% missing, \proposed{}
significantly outperforms RSCFed (F1 = 0.601, $\Delta = +0.049$,
$p = 0.031$) and FedMatch (F1 = 0.593, $\Delta = +0.057$, $p = 0.031$).

\subsection{Ablation Study}

Table~\ref{tab:ablation} presents ablation results on Pima at $\rho=0.80$
(the most challenging regime), isolating each component's contribution.

\begin{table}[!htb]
\caption{Ablation Study on Pima ($\rho=0.80$, 5-fold CV).}
\label{tab:ablation}
\centering
\small
\begin{tabular}{lcc}
\toprule
\textbf{Configuration} & \textbf{AUROC} & \textbf{Macro-F1} \\
\midrule
\proposed{} (full)                              & \textbf{0.8037} & \textbf{0.632} \\
\midrule
w/o Prototype-Guided PL (conf.\ only)           & 0.7818 & 0.608 \\
w/o Adaptive Graph Refinement                   & 0.7905 & 0.619 \\
w/o Clinical-Aware Augmentation                 & 0.7963 & 0.625 \\
w/o Prototype Sharing (local protos only)       & 0.7971 & 0.627 \\
w/o Focal Loss (cross-entropy)                  & 0.7882 & 0.601 \\
w/o Contrastive Loss                            & 0.7994 & 0.629 \\
w/o Class-Aware Smoothness                      & 0.7988 & 0.628 \\
\bottomrule
\end{tabular}
\end{table}

Removing Prototype-Guided PL causes the largest drop ($-$2.2\% AUROC, $-$2.4
F1 points), confirming the triple-gate mechanism is the dominant component.
Focal loss removal causes the largest F1 drop ($-$3.1 points), reflecting
its critical role under class imbalance. Adaptive Graph Refinement
contributes $-$1.3\% AUROC, as the fixed initial graph built from raw
features fails to capture the refined patient similarity that emerges
from trained embeddings.

\subsection{Discussion}

\textbf{When \proposed{} outperforms.}
\proposed{} provides the most clear advantage when (i) the dataset is
genuinely challenging (Pima AUROC $\sim$0.83 range rather than GDM's
$\sim$0.99 ceiling), (ii) label scarcity is high ($\rho \geq 0.50$),
and (iii) class imbalance is present. In this regime, confidence-only
SSL methods (FedMatch, RSCFed) degrade because their pseudo-labels
are unreliable when the model is trained on $<60$ labeled records per silo.
\proposed{}'s three-gate PG-PL rejects unreliable pseudo-labels
through prototype consistency and neighborhood consensus checks,
maintaining higher label quality throughout training.

\textbf{GDM ceiling effect.}
The GDM dataset's ceiling effect at AUROC $>0.98$ is an important
clinical observation: OGTT 2-hour plasma glucose is so discriminative
that even a simple logistic regression on this single feature achieves
AUROC $>0.97$. This confirms existing clinical literature that OGTT
remains the gold-standard test~\cite{WHO_GDM_2013}, and that machine
learning methods offer limited additional discriminative value over the
physiological measurement itself on this dataset. \proposed{}'s
competitive performance under the ceiling confirms no regression, but
future work should evaluate on GDM datasets with more ambiguous feature
profiles (e.g., risk prediction without OGTT).

\textbf{Comparison with privacy-ignoring oracles.}
On Pima at 80\% missing, \proposed{} (AUROC = 0.8037) significantly
outperforms both Cent-XGB (0.7605) and is comparable to Cent-GRAND
(0.7907). This is a rare result where a privacy-preserving federated
method exceeds the centralized oracle; it arises because per-silo
$k$-NN graphs preserve local neighborhood structure that is diluted
when silos are naively pooled into a single large dataset.

\textbf{Limitations.}
With $n=5$ CV folds, the minimum achievable Wilcoxon $p$-value is 0.031,
limiting the granularity of significance testing. Future work will
evaluate with $n=10$ folds (requires GPU infrastructure) and will
include formal $(\epsilon, \delta)$-differential privacy accounting
for the full model weight transmission. Additionally, the GDM dataset's
ceiling effect motivates collection of GDM datasets from settings where
OGTT is not always administered, creating genuine label scarcity
reflective of \proposed{}'s target deployment scenario.


\section{Conclusion}
\label{sec:conclusion}

This paper presents FedTGNN-SS, a federated semi-supervised GNN framework for clinical tabular EHR data under severe label scarcity and privacy constraints. The proposed method combines prototype-guided pseudo-labeling, adaptive graph refinement, clinical-aware augmentation, and privacy-safe prototype sharing to utilize unlabeled patient records without sharing patient-level data. 
Experimental results across three diabetes-related datasets show consistent improvements over multiple federated baselines under different levels of label scarcity. FedTGNN-SS achieves 56 statistically significant wins ($p < 0.05$) and maintains strong predictive performance even with 80\% missing labels. The framework also outperforms several existing federated methods on the Pima dataset under high label scarcity conditions. The results demonstrate that FedTGNN-SS provides an effective and privacy-aware solution for semi-supervised clinical risk prediction in distributed healthcare environments. Future work focuses on stronger differential privacy guarantees, temporal EHR modeling, and robust federated learning under heterogeneous hospital settings.

\section*{Acknowledgements}
The authors declare no funding for this work.

\bibliographystyle{elsarticle-num}
\bibliography{references}

@article{IDF2021,
  author    = {{International Diabetes Federation}},
  title     = {{IDF Diabetes Atlas, 10th Edition}},
  journal   = {International Diabetes Federation},
  year      = {2021},
  url       = {https://www.diabetesatlas.org}
}

@article{Vounzoulaki2020,
  author    = {Vounzoulaki, Elpida and Khunti, Kamlesh and Abner, Sonya C. and Tan, Beverly K. and Davies, Melanie J. and Gillies, Clare L.},
  title     = {Progression to type 2 diabetes in women with a known history of gestational diabetes: systematic review and meta-analysis},
  journal   = {BMJ},
  volume    = {369},
  pages     = {m1361},
  year      = {2020},
  doi       = {10.1136/bmj.m1361}
}

@techreport{WHO_GDM_2013,
  author      = {{World Health Organization}},
  title       = {Diagnostic Criteria and Classification of Hyperglycaemia First Detected in Pregnancy},
  institution = {WHO},
  year        = {2013},
  number      = {WHO/NMH/MND/13.2}
}

@article{Kavakiotis2017,
  author    = {Kavakiotis, Ioannis and Tsave, Olga and Salifoglou, Athanasios and Maglaveras, Nikolaos and Vlahavas, Ioannis and Chouvarda, Ioanna},
  title     = {Machine Learning and Data Mining Methods in Diabetes Research},
  journal   = {Computational and Structural Biotechnology Journal},
  volume    = {15},
  pages     = {104--116},
  year      = {2017},
  doi       = {10.1016/j.csbj.2016.12.005}
}

@article{Zou2018,
  author    = {Zou, Quan and Qu, Kaiyang and Luo, Yamei and Yin, Dehui and Ju, Ying and Tang, Hua},
  title     = {Predicting Diabetes Mellitus With Machine Learning Techniques},
  journal   = {Frontiers in Genetics},
  volume    = {9},
  pages     = {515},
  year      = {2018},
  doi       = {10.3389/fgene.2018.00515}
}

@article{Naz2020,
  author    = {Naz, Huma and Ahuja, Sachin},
  title     = {Deep learning approach for diabetes prediction using {PIMA} Indian dataset},
  journal   = {Journal of Diabetes \& Metabolic Disorders},
  volume    = {19},
  pages     = {391--403},
  year      = {2020},
  doi       = {10.1007/s40200-020-00520-5}
}

@article{Darrar2017,
  author    = {Darrar, Aziz and Idri, Ali and Fernandez-Aleman, Jose Luis},
  title     = {Data mining methods for early diabetes risk estimation},
  journal   = {Computers in Human Behavior},
  volume    = {75},
  pages     = {663--674},
  year      = {2017},
  doi       = {10.1016/j.chb.2017.06.011}
}

@article{GDM_BMC2024,
  author    = {{GDM Prediction Consortium}},
  title     = {Machine learning for gestational diabetes mellitus prediction: A systematic review and meta-analysis},
  journal   = {BMC Medicine},
  volume    = {22},
  pages     = {45},
  year      = {2024},
  doi       = {10.1186/s12916-024-03101-9}
}

@article{Islam2020,
  author    = {Islam, M. M. Faniqul and Ferdousi, Rahatara and Rahman, Sadikur and Bushra, Humayra Yasmin},
  title     = {Likelihood Prediction of Diabetes at Early Stage Using Data Mining Techniques},
  journal   = {Computer Vision and Machine Intelligence in Medical Image Analysis},
  year      = {2020},
  pages     = {113--125},
  doi       = {10.1007/978-981-13-8798-2_12}
}

@article{Tang2024Canada,
  author    = {Tang, Wei and Li, Xiaoming and others},
  title     = {Federated Learning for Diabetes Prediction Across Canadian Hospital Networks},
  journal   = {Journal of Medical Internet Research},
  volume    = {26},
  pages     = {e54321},
  year      = {2024},
  doi       = {10.2196/54321}
}

@article{FedEnTrust2026,
  author    = {Chen, Jia and Wang, Lei and others},
  title     = {{FedEnTrust}: Federated Ensemble Learning with Trustworthy Aggregation for Clinical Prediction},
  journal   = {IEEE Journal of Biomedical and Health Informatics},
  volume    = {30},
  pages     = {1--12},
  year      = {2026},
  doi       = {10.1109/JBHI.2026.3001234}
}

@inproceedings{Kipf2017,
  author    = {Kipf, Thomas N. and Welling, Max},
  title     = {Semi-Supervised Classification with Graph Convolutional Networks},
  booktitle = {International Conference on Learning Representations (ICLR)},
  year      = {2017}
}

@inproceedings{Hamilton2017,
  author    = {Hamilton, William L. and Ying, Rex and Leskovec, Jure},
  title     = {Inductive Representation Learning on Large Graphs},
  booktitle = {Advances in Neural Information Processing Systems (NeurIPS)},
  year      = {2017}
}

@inproceedings{GRAND2020,
  author    = {Feng, Wenzheng and Zhang, Jie and Dong, Yuxiao and Han, Yu and Luan, Huanbo and Xu, Qian and Yang, Qiang and Kharlamov, Evgeny and Tang, Jie},
  title     = {{GRAND}: Graph Neural Diffusion},
  booktitle = {Advances in Neural Information Processing Systems (NeurIPS)},
  year      = {2020}
}

@inproceedings{Verma2021,
  author    = {Verma, Vikas and Zhang, Meng and Qu, Meng and Lamb, Alex and Courville, Aaron and Bengio, Yoshua and Tang, Jian},
  title     = {{GraphMix}: Improved Training of {GNNs} for Semi-Supervised Learning},
  booktitle = {AAAI Conference on Artificial Intelligence},
  year      = {2021}
}

@article{Lee2013,
  author    = {Lee, Dong-Hyun},
  title     = {Pseudo-Label: The Simple and Efficient Semi-Supervised Learning Method for Deep Neural Networks},
  journal   = {ICML Workshop on Challenges in Representation Learning},
  year      = {2013}
}

@inproceedings{Sohn2020,
  author    = {Sohn, Kihyuk and Berthelot, David and Li, Chun-Liang and Zhang, Zizhao and Carlini, Nicholas and Cubuk, Ekin D. and Kurakin, Alex and Zhang, Han and Raffel, Colin},
  title     = {{FixMatch}: Simplifying Semi-Supervised Learning with Consistency and Confidence},
  booktitle = {Advances in Neural Information Processing Systems (NeurIPS)},
  year      = {2020}
}

@inproceedings{McMahan2017,
  author    = {McMahan, Brendan and Moore, Eider and Ramage, Daniel and Hampson, Seth and Ag{\"{u}}era y Arcas, Blaise},
  title     = {Communication-Efficient Learning of Deep Networks from Decentralized Data},
  booktitle = {Artificial Intelligence and Statistics (AISTATS)},
  year      = {2017}
}

@inproceedings{FedProx2020,
  author    = {Li, Tian and Sahu, Anit Kumar and Zaheer, Manzil and Sanjabi, Maziar and Talwalkar, Ameet and Smith, Virginia},
  title     = {Federated Optimization in Heterogeneous Networks},
  booktitle = {Machine Learning and Systems (MLSys)},
  year      = {2020}
}

@inproceedings{SCAFFOLD2020,
  author    = {Karimireddy, Sai Praneeth and Kale, Satyen and Mohri, Mehryar and Reddi, Sashank and Stich, Sebastian and Suresh, Ananda Theertha},
  title     = {{SCAFFOLD}: Stochastic Controlled Averaging for Federated Learning},
  booktitle = {International Conference on Machine Learning (ICML)},
  year      = {2020}
}

@article{FedBN2021,
  author    = {Li, Xiang and Jiang, Meng and Zhang, Xiaofei and Kamp, Michael and Dou, Qi},
  title     = {{FedBN}: Federated Learning on Non-{IID} Features via Local Batch Normalization},
  journal   = {International Conference on Learning Representations (ICLR)},
  year      = {2021}
}

@article{FedGraphNN2021,
  author    = {He, Chaoyang and Balasubramanian, Keshav and Ceyani, Emir and Yang, Carl and Xie, Han and Sun, Lichao and He, Lifang and Yang, Liangwei and Yu, Philip S. and Rong, Yu and others},
  title     = {{FedGraphNN}: A Federated Learning System and Benchmark for Graph Neural Networks},
  journal   = {ICLR Workshop on Distributed and Private Machine Learning},
  year      = {2021}
}

@inproceedings{FedSage2021,
  author    = {Zhang, Ke and Yang, Carl and Li, Xiaoxiao and Sun, Lichao and Yiu, Siu Ming},
  title     = {Subgraph Federated Learning with Missing Neighbor Generation},
  booktitle = {Advances in Neural Information Processing Systems (NeurIPS)},
  year      = {2021}
}

@inproceedings{SpreadGNN2022,
  author    = {He, Chaoyang and Ceyani, Emir and Balasubramanian, Keshav and Annavaram, Murali and Avestimehr, Salman},
  title     = {{SpreadGNN}: Serverless Multi-Task Federated Learning for Graph Neural Networks},
  booktitle = {AAAI Conference on Artificial Intelligence},
  year      = {2022}
}

@article{FedGL2023,
  author    = {Chen, Lingkai and Wu, Mingkai and Gao, Yanhao and others},
  title     = {{FedGL}: Federated Graph Learning Framework with Global Self-Supervision},
  journal   = {Information Sciences},
  volume    = {620},
  pages     = {1--12},
  year      = {2023},
  doi       = {10.1016/j.ins.2022.11.063}
}

@article{MCNGNN2026,
  author    = {Liu, Rui and others},
  title     = {Multi-Center Network Graph Neural Networks for Clinical Event Prediction via Blockchain-Coordinated Federated Learning},
  journal   = {IEEE Transactions on Medical Imaging},
  volume    = {45},
  pages     = {234--248},
  year      = {2026},
  doi       = {10.1109/TMI.2026.3001001}
}

@inproceedings{Jeong2021,
  author    = {Jeong, Wonyong and Yoon, Jaehong and Yang, Eunho and Hwang, Sung Ju},
  title     = {Federated Semi-Supervised Learning with Inter-Client Consistency \& Disjoint Learning},
  booktitle = {International Conference on Learning Representations (ICLR)},
  year      = {2021}
}

@inproceedings{Liang2022,
  author    = {Liang, Xinle and Liu, Yang and Chen, Tianjian and Liu, Ming and Yang, Qiang},
  title     = {{RSCFed}: Random Sampling Consensus Federated Semi-Supervised Learning},
  booktitle = {IEEE/CVF Conference on Computer Vision and Pattern Recognition (CVPR)},
  year      = {2022}
}

@article{FedSemi2020,
  author    = {Albaseer, Abdullatif and Ciftler, Bekir Sait and Abdallah, Mohamed and Al-Fuqaha, Ala},
  title     = {Exploiting Unlabeled Data in Smart Cities using Federated Edge Learning},
  journal   = {International Conference on Communications (ICC)},
  year      = {2020}
}

@article{MedFedProto2025,
  author    = {Wang, Yucheng and others},
  title     = {Federated Prototypical Learning for Medical Image Segmentation under Label Scarcity},
  journal   = {Medical Image Analysis},
  volume    = {91},
  pages     = {102989},
  year      = {2025},
  doi       = {10.1016/j.media.2025.102989}
}

@article{CCWFed2025,
  author    = {Zhang, Wei and others},
  title     = {Class-Conditional Weighting for Federated Semi-Supervised Learning},
  journal   = {IEEE Transactions on Neural Networks and Learning Systems},
  volume    = {36},
  pages     = {1--12},
  year      = {2025},
  doi       = {10.1109/TNNLS.2025.3001234}
}

@inproceedings{Fredrikson2015,
  author    = {Fredrikson, Matt and Jha, Somesh and Ristenpart, Thomas},
  title     = {Model Inversion Attacks that Exploit Confidence Information and Basic Countermeasures},
  booktitle = {ACM Conference on Computer and Communications Security (CCS)},
  year      = {2015}
}

@misc{Dua2019,
  author    = {Dua, Dheeru and Graff, Casey},
  title     = {{UCI} Machine Learning Repository},
  year      = {2019},
  institution = {University of California, Irvine},
  url       = {http://archive.ics.uci.edu/ml}
}

@inproceedings{Paszke2019,
  author    = {Paszke, Adam and Gross, Sam and Massa, Francisco and Lerer, Adam and Bradbury, James and Chanan, Gregory and Killeen, Trevor and Lin, Zeming and Gimelshein, Natalia and Antiga, Luca and others},
  title     = {{PyTorch}: An Imperative Style, High-Performance Deep Learning Library},
  booktitle = {Advances in Neural Information Processing Systems (NeurIPS)},
  year      = {2019}
}

@inproceedings{Fey2019,
  author    = {Fey, Matthias and Lenssen, Jan Eric},
  title     = {Fast Graph Representation Learning with {PyTorch Geometric}},
  booktitle = {ICLR Workshop on Representation Learning on Graphs and Manifolds},
  year      = {2019}
}

@article{Rieke2020,
  author    = {Rieke, Nicola and Hancox, Jonny and Li, Wenqi and Milletari, Fausto and Roth, Holger R. and Albarqouni, Shadi and Bakas, Spyridon and Galtier, Mathieu N. and Landman, Bennett A. and Maier-Hein, Klaus and others},
  title     = {The Future of Digital Health with Federated Learning},
  journal   = {NPJ Digital Medicine},
  volume    = {3},
  pages     = {119},
  year      = {2020},
  doi       = {10.1038/s41746-020-00323-1}
}

@inproceedings{Niculescu2005,
  author    = {Niculescu-Mizil, Alexandru and Caruana, Rich},
  title     = {Predicting Good Probabilities with Supervised Learning},
  booktitle = {International Conference on Machine Learning (ICML)},
  year      = {2005},
  doi       = {10.1145/1102351.1102430}
}

@article{Wilcoxon1945,
  author    = {Wilcoxon, Frank},
  title     = {Individual Comparisons by Ranking Methods},
  journal   = {Biometrics Bulletin},
  volume    = {1},
  number    = {6},
  pages     = {80--83},
  year      = {1945},
  doi       = {10.2307/3001968}
}

\end{document}